\documentclass[letterpaper, 10 pt, conference]{ieeeconf}

\IEEEoverridecommandlockouts                          
\usepackage{amsmath}
\usepackage{amssymb} 
\usepackage[usenames, dvipsnames, table]{xcolor}
\usepackage[noadjust]{cite}
\usepackage{breqn}
\usepackage{overpic}
\usepackage{svg}

\usepackage[small]{caption}
\usepackage{subcaption}

\usepackage{tikz}
\usepackage{textcomp}
\usepackage{hyperref}
\usepackage{lipsum}
\usepackage{algorithm}
\usepackage{algpseudocode}

\newcommand\copyrighttext{
  \footnotesize \textcopyright 2022 IEEE. Personal use of this material is permitted.
  Permission from IEEE must be obtained for all other uses, in any current or future
  media, including reprinting/republishing this material for advertising or promotional
  purposes, creating new collective works, for resale or redistribution to servers or
  lists, or reuse of any copyrighted component of this work in other works.}
\newcommand\copyrightnotice{%
\begin{tikzpicture}[remember picture,overlay]
\node[anchor=south,yshift=10pt] at (current page.south) {\fbox{\parbox{\dimexpr\textwidth-\fboxsep-\fboxrule\relax}{\copyrighttext}}};
\end{tikzpicture}%
}

\usepackage[markup=bfit, deletedmarkup=sout, authormarkup=superscript,commandnameprefix=always]{changes}
\definechangesauthor[name={ar}, color=teal]{AR}
\definechangesauthor[name={NN}, color=MidnightBlue]{NN}
\definechangesauthor[name={ce}, color=ForestGreen]{CE}
\definechangesauthor[name={ms}, color=red]{MS}
\definechangesauthor[name={sk}, color=purple]{SK}

\makeatletter
\let\NAT@parse\undefined
\makeatother

\usepackage{hyperref}
\usepackage[capitalize]{cleveref}

\hypersetup{
  linkcolor  = BrickRed,
  citecolor  = Green,
  urlcolor   = NavyBlue,
  colorlinks = true,
}

\hypersetup{
  linkcolor  = black,
  citecolor  = black,
  urlcolor   = black,
  colorlinks = true,
}

\title{\LARGE \bf A Framework for the Systematic Evaluation of Obstacle Avoidance and Object-Aware Controllers}

\author{Caleb Escobedo$^{*}$, Nataliya Nechyporenko, Shreyas Kadekodi, Alessandro Roncone %
\thanks{* Corresponding author.}
\thanks{This work was supported by the Office of Naval Research under grant N00014-22-1-2482. NN was supported by NSF DGE 2040434.}
\thanks{All authors are with the Department of Computer Science, University of Colorado Boulder, 1111 Engineering Drive, Boulder, CO USA {\tt\small name.surname@colorado.edu}}
\thanks{This paper was published at the IEEE/RSJ International Conference on Intelligent Robots and Systems (IROS) 2022. 
DOI: \href{https://ieeexplore.ieee.org/document/9982198}{0.1109/IROS47612.2022.9982198}}
}

\begin{document}
\maketitle
\thispagestyle{empty}
\pagestyle{empty}
\copyrightnotice
\begin{abstract}
Real-time control is an essential aspect of safe robot operation in the real world with dynamic objects. We present a framework for the analysis of object-aware controllers, methods for altering a robot's motion to anticipate and avoid possible collisions. This framework is focused on three design considerations: kinematics, motion profiles, and virtual constraints. Additionally, the analysis in this work relies on verification of robot behaviors using fundamental robot-obstacle experimental scenarios. To showcase the effectiveness of our method we compare three representative object-aware controllers. The comparison uses metrics originating from the design considerations. From the analysis, we find that the design of object-aware controllers often lacks kinematic considerations, continuity of control points, and stability in movement profiles. We conclude that this framework can be used in the future to design, compare, and benchmark obstacle avoidance methods.
\end{abstract}
\section{Introduction}\label{sec:intro}
Ensuring safety is paramount as robots transition into spaces occupied by people in highly dynamic environments such as hospitals, homes, and schools \cite{zacharaki2020safety}. 
Safety implies the ability to perceive and act on information in real-time.
Multiple strategies have been introduced to help robots avoid harmful contact and achieve physical safety \cite{lasota2017survey}. 
Unlike control methods, complex motion planners and prediction models cannot satisfy real-time safety requirements due to their computational complexity.
Controllers can react on-line to obstacles in their environment by changing the robot's motion to avoid or anticipate contact.
We define this category of controllers as object-aware controllers (OACs). OACs are used to avoid and reduce the force of contact with obstacles not accounted for by a robot's trajectory planning system. Scenarios such as this can be due to errors in perception, occlusions, dynamic obstacles, or presence of high clutter.  
Despite the advantages of control based methods, limited analysis has been conducted on the strengths and limitations of OACs \cite{lasota2017survey}. Without analysis, it is difficult to compare existing controllers or propose actionable insights for future direction of research.

In this work, we design and test a systematic approach to object aware controller analysis. The novelty of our work lies in (1) defining fundamental design considerations for structured and robust analysis of OACs in varied contexts; (2) comparing three representative OACs based on our criteria to highlight their strengths and weaknesses; and (3) proposing directions for future research in the field based on our findings. This analysis will bridge the gap between the understanding of robot movement and the design of algorithms that result in a particular motion. This is an essential part of building intelligent systems that work in complex, dynamic environments. 

\begin{figure}
    \centering
    \includegraphics[width=1\linewidth, height=5.5cm]{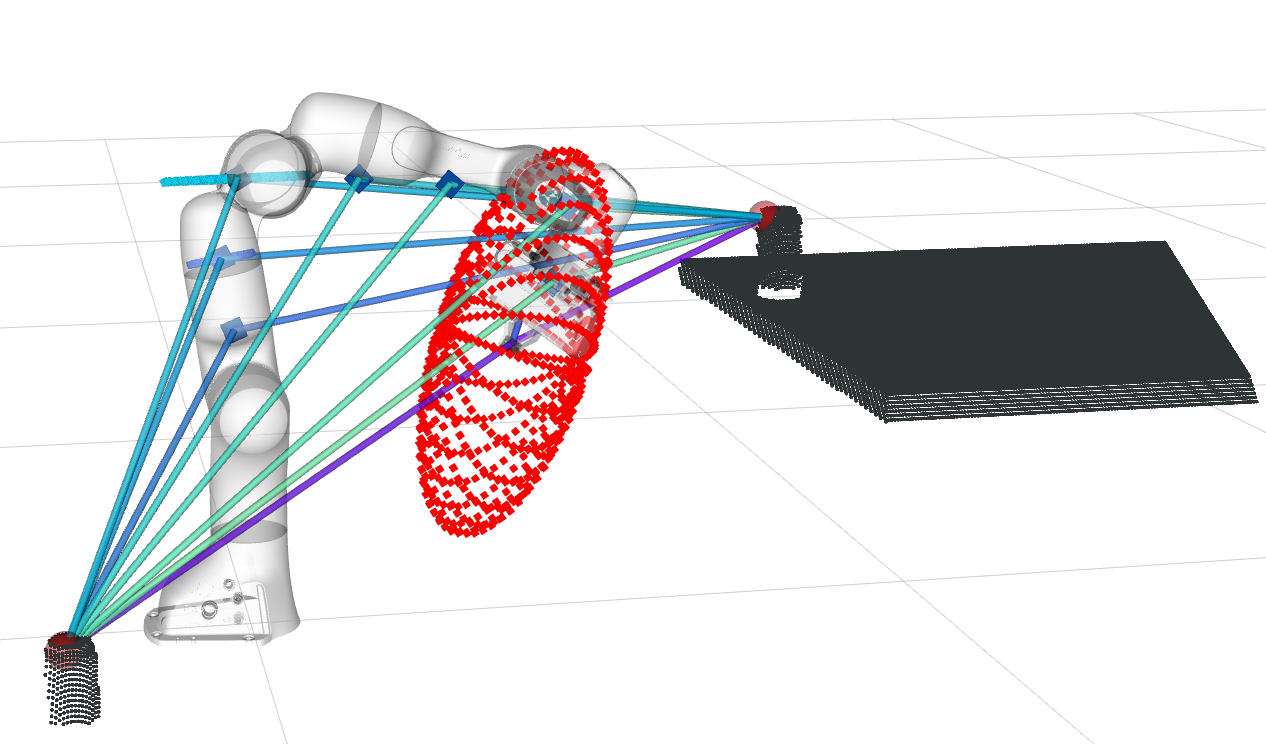}
        \caption{This diagram depicts a scenario involving a robotic manipulator and two dynamic obstacles, shown as small black objects. The colored rays emanating from these obstacles represent the distance from the obstacles to various control points along the robot's body. The red ellipsoid  illustrates the current range of motion for the end-effector. The ellipsoid and rays are components of the analysis of object aware controllers presented in this work.}
    \label{fig:rviz}\vspace{-18pt}
\end{figure}
\begin{figure*}
    \centering
        \begin{minipage}{.3\textwidth}
            \begin{subfigure}{\textwidth}
            \centering
            \includegraphics[width=\textwidth, height = 4.8cm]{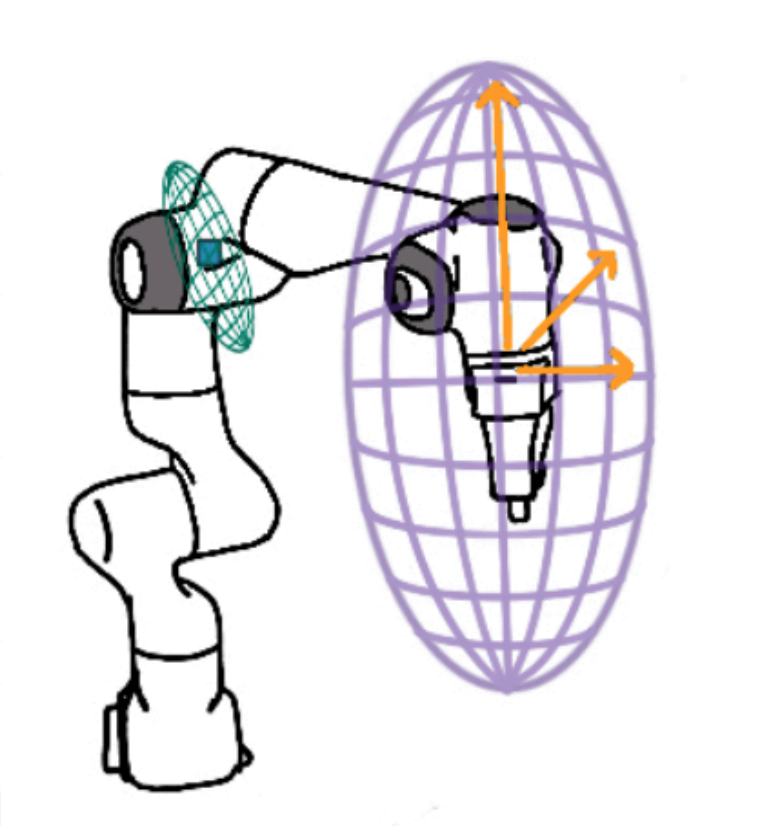}
            \caption{High Manipulability}\label{fig:analysis_ellipse_1_high}
            \end{subfigure}\\
            \begin{subfigure}{\textwidth}
            \centering
            \includegraphics[width=\textwidth, height = 4.8cm]{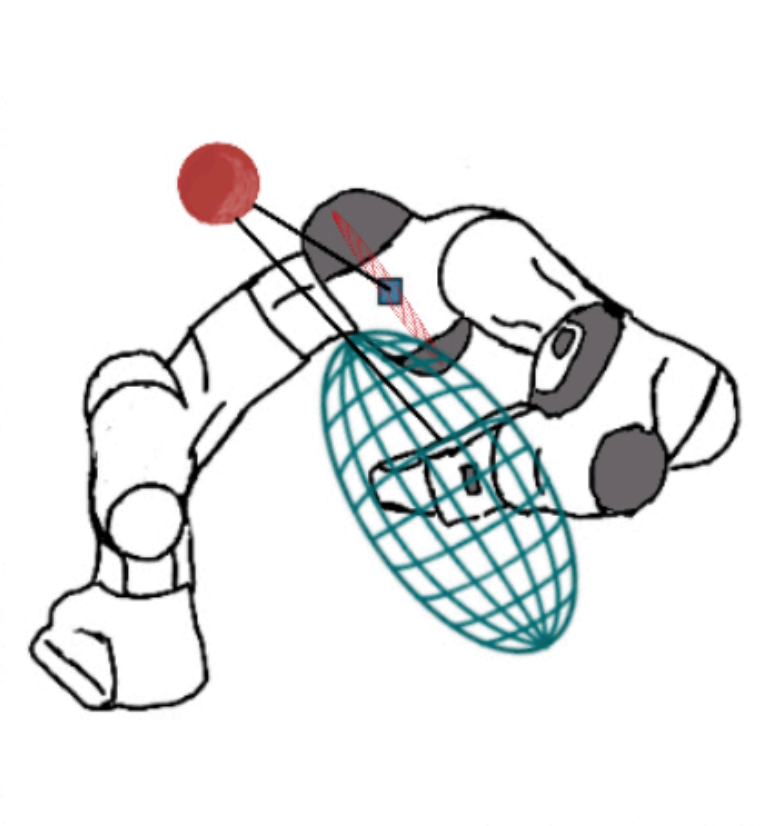}
            \caption{Low Manipulability}\label{fig:analysis_ellipse_2_low}
            \end{subfigure}
        \end{minipage}
        \hfill
        \begin{minipage}{.3\textwidth}
            \begin{subfigure}{\textwidth}
            \centering
            \includegraphics[width=\textwidth, height = 4.8cm]{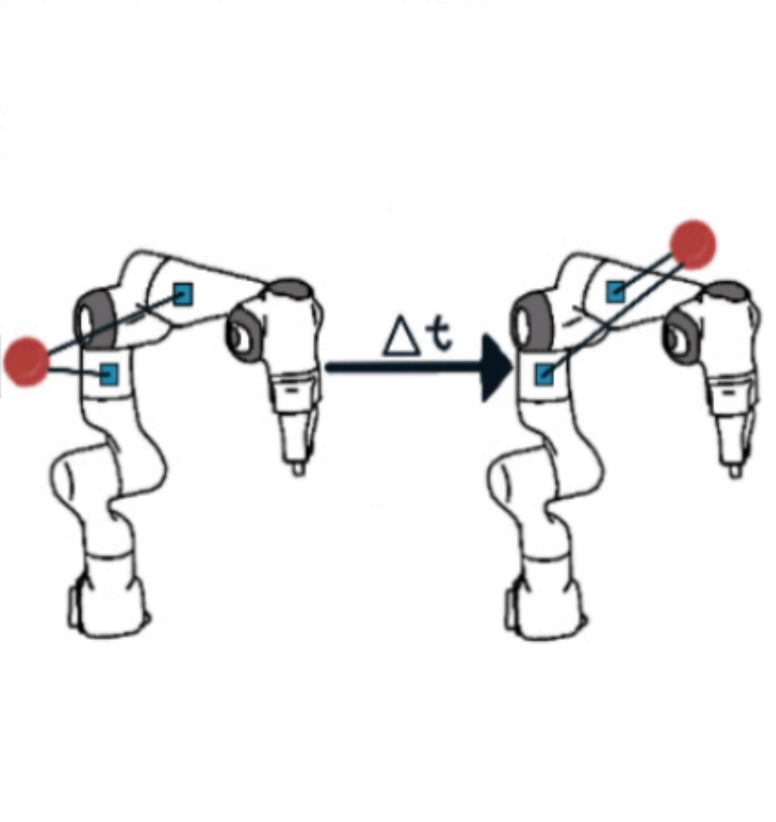}
            \caption{Static Control Points}\label{fig:analysis_control_points_1_static}
            \end{subfigure}\\
            \begin{subfigure}{\textwidth}
            \centering
            \includegraphics[width=\textwidth, height = 4.8cm]{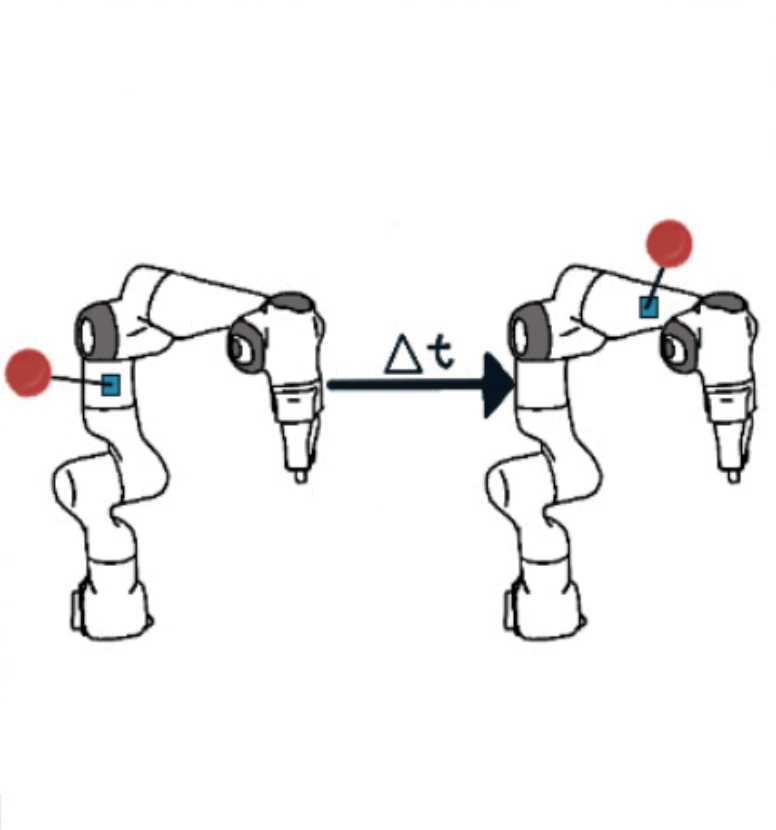}
            \caption{Dynamic Control Points}\label{fig:analysis_control_points_2_dynamic}
            \end{subfigure}
        \end{minipage}
        \hfill
        \begin{minipage}{.3\textwidth}
            \begin{subfigure}{\textwidth}
            \centering
            \includegraphics[width=\textwidth, height = 4.8cm]{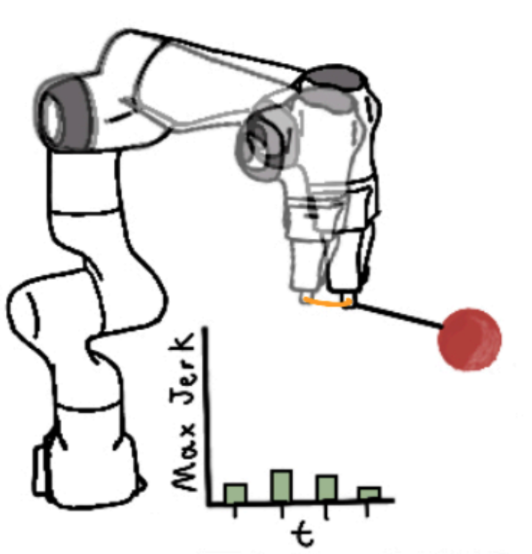}
            \caption{Low jerk}\label{fig:analysis_jerk_1_low}
            \end{subfigure}\\
            \begin{subfigure}{\textwidth}
            \centering
            \includegraphics[width=\textwidth, height = 4.8cm]{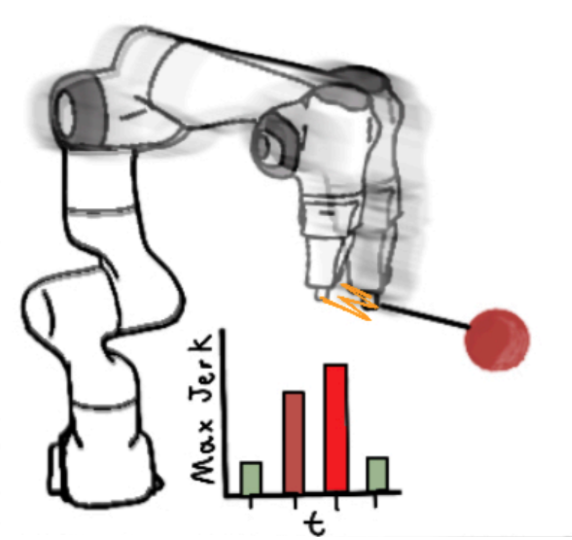}
            \caption{High jerk}\label{fig:analysis_jerk_2_high}
            \end{subfigure}
        \end{minipage}
        \caption{ Each column represents one of the three design considerations used to evaluate controllers in this work, introduced in \Cref{sec:evaluation}.
        \textbf{a)} High-manipulability configuration as shown from the large ellipsoids. Eigenvectors of $\mathbf{J J}^{\top}$, where $\mathbf{J}$ is the robot Jacobian introduced in \Cref{sec:evaluation}, are shown as orange arrows. 
        \textbf{b)} An undesirable configuration due to movement restrictions caused by the red object, the smaller the ellipsoids are, the less ability the robot has to move a particular point freely. 
        \textbf{c)} Static control points are determined by the controller designer and placed in predefined locations. Used in \cite{flacco2015control}. 
        \textbf{d)} Dynamic control points are selected by determining the closest point on the robot body to an obstacle. Used in \cite{ding2020collision} and \cite{escobedo2021contact}. 
        \textbf{e)} Low jerk is exhibited and the robot smoothly moves away from the obstacle as seen by the EE orange path. 
        \textbf{f)} The movement restrictions imposed on the EE cause the jittery motion as seen in the jagged path to its final location.}
        \vspace{-18pt}
\end{figure*}
The experiments presented in this work are based on fundamental robot-obstacle interactions. For example, \Cref{fig:rviz} shows a snapshot of an avoidance interaction between an obstacle and robot. The red dotted ellipsoid containing the robot end-effector (EE) is a representation of the robot's manipulability (see \Cref{sec:evaluation}) in the current joint configuration. We track the resulting movement of several avoidance methods. This information is then distilled into a set of understandable metrics and graphs for each control strategy. Evaluation criteria, used in this work to examine three OACs (\cite{flacco2015control}, \cite{ding2020collision}, \cite{escobedo2021contact}), is developed based on the following design considerations:
1) Kinematics - OACs are analyzed based on their capacity to take into account the position and movement of multi-degree of freedom kinematic chains that form the structure of a robotic manipulator;
2) Motion profiles - OACs are analyzed based on their ability to produce a smooth velocity, acceleration, jerk, and end-effector path profile;
3) Virtual constraints - OACs are analyzed based on their ability to generate movement requirements in response to their environment. By considering kinematics, motion profiles, and virtual constraints, we assess three representative OACs and conclude that our approach can be used to showcase the reactive motion of future OACs, in addition to creating a baseline for comparison with current methods. 

This paper is organized as follows: \cref{sec:related_work} provides background on OACs;
\cref{sec:methods} describes each controller and its avoidance mechanism in detail; \cref{sec:evaluation} introduces our evaluation metrics based on our design criteria; \cref{sec:methods:experimental_setp} outlines our experimental setup, and \cref{sec:results_discussion} presents our results and conclusions.
\section{Related Work}\label{sec:related_work}
Robotic manipulation in dynamic environments requires real-time trajectory adjustment to avoid or anticipate collisions while continuing to execute a task. Historically, to ensure safe interaction with no external environmental sensors, collision detection has been used to quickly stop a manipulator \cite{haddadin2017robot}. This method does reduce the overall force and increase safety during interactions, but the robot stops movement suddenly and cannot anticipate or avoid contact.
Various OACs allow robot manipulators to avoid collisions with dynamic obstacles in real-time \cite{escobedo2021contact, ding2020collision, flacco2012depth, rakita2021collisionik, merckaert2022real, fan2021aurasense, nguyen2018compact, roncone2015learning}. An early method for obstacle avoidance \cite{flacco2012depth} uses
repulsive vectors, a simplified form of artificial potential fields \cite{khatib1986real}, to alter the robot's end-effector (EE) trajectory while it continues towards the robot's main task. An alternative avoidance method using an Explicit Reference Governor formalism was used to ensure that particular safety constraints between a robot manipulator and human are not violated \cite{merckaert2022real}.
Our previous work enabled robot manipulators to anticipate contact, while still attempting to avoid collisions when possible \cite{escobedo2021contact}. Nearby obstacle position information has also been used to alter impedance controlled robot stiffness values \cite{ding2021improving} to soften collisions. Lastly, the per-instant pose optimization method CollisionIK introduced in \cite{rakita2021collisionik} uses a normalized objective function introduced in \cite{rakita2018relaxedik} to weight its avoidance terms. 

While all of the avoidance methods introduced in this section allow for some variation of obstacle avoidance, there exists a significant gap in knowledge of limitations, benefits, and practical use of each in real-world scenarios. Through analysis of existing OACs, we outline several criteria for evaluating and comparing their real-world performance.

\section{Object-Aware Controllers (OACs)}\label{sec:methods}
In this work, we analyze characteristics of three Object Aware Controllers (OACs). We use the primary author's last name to refer to each controller throughout this work. For example, the controller introduced in \cite{flacco2012depth} will be referred to as ``Flacco''. The other two methods used in this work are ``Ding'' \cite{ding2020collision} and ``Escobedo'' \cite{escobedo2021contact}. The controllers we evaluate in this work are chosen due to their similarities in optimization formulation and applied restrictions.
In this section, we formalize each controller's specific avoidance mechanisms. 

\subsection{Assumptions for the evaluation of OACs} We introduce a general obstacle representation where we assume that a robot's perception system can reduce any pertinent sensor data to a finite set of rigid obstacles $\boldsymbol{O}$, where each element $\boldsymbol{o} \in\boldsymbol{O}$ contains all information required for a particular controller. Each of the evaluated OAC's main tasks and avoidance constraints are expressed in quadratic programming (QP) notation, introduced in \Cref{eq:qp}. 
\begin{equation}\label{eq:qp}\small
\min _{\dot{\mathbf{q}}} \frac{1}{2} \dot{\mathbf{q}}^{\top} \mathbf{H} \dot{\mathbf{q}}+f^{\top} \dot{\mathbf{q}} \quad \text { s.t. }\left\{\begin{array}{l}
\mathbf{A} \dot{\mathbf{q}} \leq \mathbf{b} \\
\mathbf{b}_{1} \leq \dot{\mathbf{q}} \leq \mathbf{b}_{\mathrm{u}}
\end{array}\right.
\end{equation}
Here $\mathbf{H}$ is the quadratic objective term and $f$ is the linear objective term created by augmenting each controller's main task, which all share the first term \Cref{eq:standard_minimization}, to fit the QP formulation. Linear inequality constraints are added in $\mathbf{A} \dot{\mathbf{q}} \leq \mathbf{b}$. Joint velocity $\mathbf{\dot{q}}$, lower limit $\mathbf{b}_{l} $, and upper limit $\mathbf{b}_{u}$ are specified in the form $\mathbf{b}_{l} \leq \dot{\mathbf{q}} \leq \mathbf{b}_{{u}}$, which for all controllers in this work is expressed:
\begin{equation} \label{eq:joint_bounds}\small
\left.\begin{array}{ll}
0, & \mathbf{q} \leq \mathbf{q}_{{l}} \\
\dot{\mathbf{q}}_{{l}}, & \text { otherwise }
\end{array}\right\}   \leq \dot{\mathbf{q}} \leq \begin{cases}0, & \mathbf{q} \geq \mathbf{q}_{{u}} \\
\dot{\mathbf{q}}_{{u}}, & \text { otherwise. }\end{cases}
\end{equation}
In \Cref{eq:joint_bounds}, $\mathbf{q}_{{l}}$ and $\dot{\mathbf{q}}_{{l}}$ represent the lower bounds of the joint position and velocity, whereas $\mathbf{q}_{{u}}$ and $\dot{\mathbf{q}}_{{u}}$ represent the upper bounds. 
All controllers evaluated in this work are implemented using Cartesian velocity control and share the same first main task term, expressed as: 
\begin{equation} \label{eq:standard_minimization}\small
\begin{split}
g(\mathbf{\dot{q})} & = \frac{1}{2}\mathbf{(\dot{x}_d - J\dot{q})^{\top}(\dot{x}_d - J\dot{q}) }
\end{split}
\end{equation}
Here, $g(\mathbf{\dot{q})} $ represents the quadratic error between the desired velocity $\mathbf{\dot{x}_d}\in \mathbb{R}^m$ and the robot's actual velocity $\mathbf{J\dot{q}}\in \mathbb{R}^m$. $\mathbf{\dot{q}} \in \mathbb{R}^n$ is the joint velocities of an $n$-joint, kinematically redundant robot manipulator. $\mathbf{J} \in \mathbb{R}^{m \times n}$ contains the first order partial-derivatives of the robot's joint positions in relation to the EE Cartesian velocity $\mathbf{\dot{x}}$. This relationship is expressed as:
\begin{equation}\label{eq:main_jacobian}\small
\mathbf{\dot{x} = J\dot{q}}
\end{equation}
\noindent Ding and Escobedo were originally introduced using a QP formulation, which allows hard constraints to be placed on manipulator motion. We implement Flacco in the QP formulation, but do not include linear inequality constraints.
\begin{figure}
    \centering
    \includegraphics[width=0.6\linewidth, height=5cm]{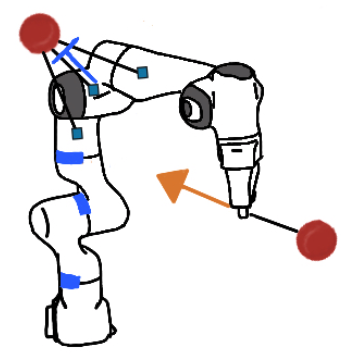}
       \put(-70,50){\color{black}$\boldsymbol{V}(\boldsymbol{p})$}
       \put(-115,129){\color{black}$\boldsymbol{J}_{p}^{T}\hat{\boldsymbol{d}}_{\boldsymbol{p}, \mathbf{o}_{\min}} \boldsymbol{r}$}
        \caption{Flacco \cite{flacco2012depth} controller diagram with obstacles represented as red circles. The orange arrow originating from the end-effector is introduced in \Cref{eq:repulsive_vector} as a repulsive force. Blue rectangles show joint restrictions caused by constraints added through \Cref{eq:flacco_body}.}
    \label{fig:flacco_controller}\vspace{-18pt}
\end{figure}
\subsection{Flacco \cite{flacco2012depth}} \label{sec:methods:flacco}
\cite{flacco2012depth} uses repulsive vectors and Cartesian constraints with artificial forces to avoid collisions.
Repulsive vectors apply a virtual force that alters the robot's end-effector trajectory in real-time. As detailed in \Cref{fig:flacco_controller}, $\boldsymbol{V}(\boldsymbol{p})$ is the repulsive vector applied to a control point $\boldsymbol{p}$. The applied repulsive force, shown in \Cref{fig:flacco_controller}, is computed as:
\begin{equation} \label{eq:repulsive_vector}\small
\begin{aligned}
\boldsymbol{V}(\boldsymbol{p}) &=v\left(\boldsymbol{p}, \boldsymbol{o}_{\min }\right)
\hat{\boldsymbol{V_{a}}}(\boldsymbol{p})
\end{aligned}
\end{equation}
Where $v(\boldsymbol{p}, \boldsymbol{o})$, the magnitude of the repulsive force from the closest obstacle $\boldsymbol{o}_{\min}$ to $\boldsymbol{p}$, is defined as: 
\begin{equation} \label{eq:repulsive magnitude}\small
v(\boldsymbol{p}, \mathbf{o})=\frac{V_{\max }}{1+e^{(\|\boldsymbol{d}_{\boldsymbol{p} \boldsymbol{o}}\|(2 / \rho)-1) \alpha}}
\end{equation} 
Here, $V_{\max }$ is a user-defined maximum velocity, $\rho$ represents the distance where the repulsive vector becomes negligible, $\boldsymbol{d}_{\boldsymbol{p}, \boldsymbol{o}}$ is the direction vector from $\boldsymbol{p}$ to $\boldsymbol{o}$, $\|\boldsymbol{d}_{\boldsymbol{p}, \boldsymbol{o}}\|$ is the distance between the two points, and $\alpha$ is a shape factor.
In \Cref{eq:repulsive_vector}
$\hat{\boldsymbol{V}}_{\boldsymbol{a}}(\boldsymbol{p})$ 
is the unit vector of the sum of all obstacles within the region of surveillance ${\boldsymbol{S}}$, defined as:
\begin{equation} \label{eq:repulsive_vector_summation}\small
\begin{aligned}
\boldsymbol{V_{a}}(\boldsymbol{p}) &=\sum_{\mathbf{o} \in \boldsymbol{S}}v (\boldsymbol{p}, \boldsymbol{o}) \hat{\boldsymbol{d}}_{\boldsymbol{p}, \boldsymbol{o}} \\
\end{aligned}
\end{equation} 
As a result, the magnitude of movement in \Cref{eq:repulsive_vector} depends on the closest obstacle point, while the direction of movement is affected by all obstacles within the region of surveillance. 

In addition, joint velocity constraints are added to avoid collisions with obstacles near the robot's body for multiple control points. For example, in \Cref{fig:flacco_controller} the blue squares are the control points where movement constraints are placed due to the nearby object. The position of a control point $\boldsymbol{p}$ is statically defined for this controller as a position along the robot's kinematic chain, as seen in \Cref{fig:analysis_control_points_1_static}. \Cref{fig:flacco_controller} shows joint restrictions being used to avoid a possible collision. 
    For the joint velocity constraints, we first calculate the risk of collision with the nearest obstacle: 
    \begin{equation}\small
        \boldsymbol{r} = \frac{v(\boldsymbol{p},\boldsymbol{o}_{\min})}{V_{\max}}
    \end{equation}
The value of $\boldsymbol{r}$ is then used to calculate the degree of influence of the constraint on each joint $\boldsymbol{s}_{i}\in \boldsymbol{s}$ where $\boldsymbol{s}$ is defined:
 \begin{equation} \small
    \boldsymbol{s}=\boldsymbol{J}_{p}^{T}\hat{\boldsymbol{d}}_{\boldsymbol{p}, \mathbf{o}_{\min}} \boldsymbol{r}
    \label{eq:flacco_body_1}
\end{equation}
where $\boldsymbol{J}_{p}$ is the Jacobian of control point $\boldsymbol{p}$. The restrictions from \Cref{eq:flacco_body_1} are shows in blue in \Cref{fig:flacco_controller}. The acceptable limits of the velocity of all joints are then set as: 
\begin{equation}\label{eq:flacco_body}\small
    \begin{array}{ll}
\text { if } \boldsymbol{s}_{i} \geq 0, & \dot{\mathbf{q}}_{\max , i}=V_{\max , i}\left(1-\boldsymbol{r})\right. \\
\text { else } & \dot{\mathbf{q}}_{\min , i}=-V_{\max , i}\left(1-\boldsymbol{r})\right.
\end{array}
\end{equation}
In situations where a moving obstacle's velocity is known, the Pivot Algorithm \cite{flacco2012depth} is used to alter the repulsive vector to move in a direction normal to the obstacle's velocity.
\\

\begin{figure}
\vspace*{0.02in}
    \centering
    \includegraphics[width=0.6\linewidth, height=4.7cm]{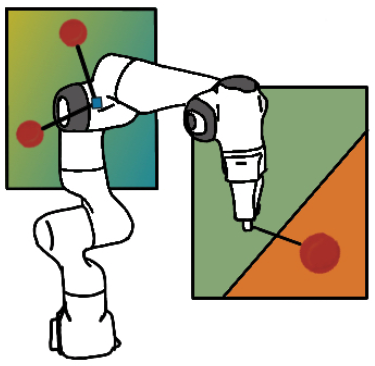}
        \put(-85, 121){\color{black}$\mathbf{w}^{\top}\nabla_{\mathbf{q}}\|\mathbf{d}\| \leq 0$}
        \put(-45,105){\color{black}$\hat{\mathbf{d}}^{\top} \mathbf{J}_{\mathrm{c}} \dot{\mathbf{q}} \leq \dot{x}_{\mathrm{a}}\quad$}
        \caption{Ding \cite{ding2020collision} controller diagram with obstacles represented as red circles. The rectangle encapsulating the EE shows movements restrictions from \Cref{eq:ding_restriction}, the orange portion shows restricted movement while the green shows where the robot can still move. This first restriction is also added to the body control points, when obstacles are nearby. The rectangle near the robot body shows the weighted sum distance gradient restriction introduce in \Cref{eq:ding_gradient}.}
\label{fig:ding_controller}
\vspace{-12pt}
\end{figure}
\subsection{Ding \cite{ding2020collision}}\label{sec:methods:ding}

 The main task equation for this controller adds a term to avoid kinematic singularities based on a robot manipulability measure, $\mu$, to \Cref{eq:standard_minimization}, expressed as:
\begin{equation} \label{eq:ding_avoid_singularities}\small
\begin{split}
g(\mathbf{\dot{q})} & = \frac{1}{2}\mathbf{(\dot{x}_d - J\dot{q})^{\top}(\dot{x}_d - J\dot{q}) } + \frac{\mu}{2}\mathbf{\dot{q}^{\top}\dot{q}}
\end{split} 
\end{equation}
see \cite{Nakamura1986} for a detailed explanation of $\mu$. The first movement constraint is added to control points along the robot body closest to each obstacle as shown in \Cref{fig:ding_controller}. In an obstacle-free scenario, the limitations would not apply and the robot would be free to move at the user-defined nominal speed.
These constraints are defined as
\begin{equation}\label{eq:ding_restriction}\small
    \hat{\mathbf{d}}^{\top} \dot{\mathbf{x}}_{\mathrm{c}}=\hat{\mathbf{d}}^{\top} \mathbf{J}_{\mathrm{c}} \dot{\mathbf{q}} \leq \dot{x}_{\mathrm{a}}\quad,
\end{equation}
where $\boldsymbol{\hat{\mathbf{d}}}$ is the unit vector of the distance between an obstacle and control point, $\dot{\mathbf{x}}_{\mathrm{c}}$ is the velocity of the control point, and $\mathbf{J}_{\mathrm{c}}$ is its Jacobian.
This additional constraint limits the approach velocity of $\dot{\mathbf{x}}_{\mathrm{c}}$ to a scalar $\dot{x}_a$, determined from a stepwise function dependent on distance to the nearest obstacle.
The second avoidance constraint, the weighted gradient of distance from $\boldsymbol{p}$ to all obstacles within a region of surveillance $\boldsymbol{o} \in \boldsymbol{S}$, is calculated as:
\begin{equation} \label{eq:ding_gradient}\small
        \mathbf{J}_{\mathrm{c}}=\mathbf{w}^{\top}\nabla_{\mathbf{q}}\|\mathbf{d}\|
\end{equation}
where $\mathbf{w}$ is a distance-based weight vector. For each obstacle, $w_{i}$ is calculated from a user-defined function. In this work we utilize the distance function from \Cref{eq:repulsive magnitude}.
 \Cref{eq:ding_gradient} causes a control point to increase the sum of all weighted distances to obstacles within the region of surveillance, visualized in \Cref{fig:ding_controller}.
The constraint added to \Cref{eq:qp} based on this term is $-\mathbf{J}_{\mathrm{c}} \dot{\mathbf{q}} \leq 0$.
\subsection{Escobedo \cite{escobedo2021contact}}\label{sec:methods:escobedo}
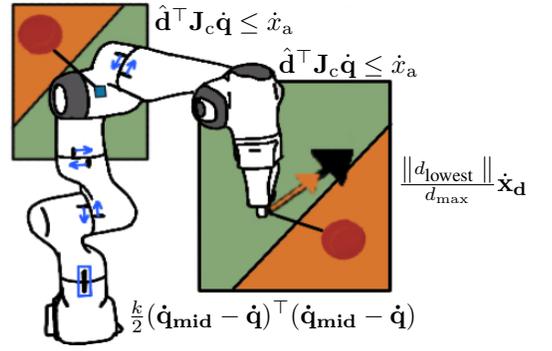
\begin{figure}
    \centering
    \vspace*{0.02in}
    \includegraphics[width=0.6\linewidth, height=4.7cm]{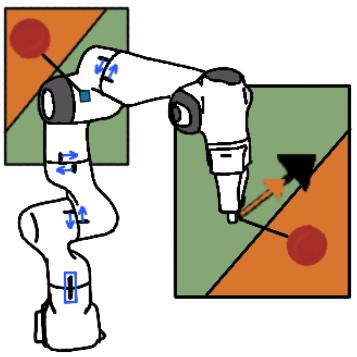}
        \put(-92, 121){\color{black}$\hat{\mathbf{d}}^{\top} \mathbf{J}_{\mathrm{c}} \dot{\mathbf{q}} \leq \dot{x}_{\mathrm{a}}\quad$}
        \put(-45,105){\color{black}$\hat{\mathbf{d}}^{\top} \mathbf{J}_{\mathrm{c}} \dot{\mathbf{q}} \leq \dot{x}_{\mathrm{a}}\quad$}
        \put(0,60){\color{black}{$\frac{\left\|d_{\text {lowest }}\right\|}{d_{\max }} \dot{\mathbf{x}}_{\mathbf{d}}$}}
        \put(-102,10){\color{black}$\frac{k}{2}(\mathbf{\dot{q}_{mid}} - \mathbf{\dot{q}})^{\top}(\mathbf{\dot{q}_{mid}} - \mathbf{\dot{q}}) \quad$}
        \caption{Escobedo \cite{escobedo2021contact} controller diagram with obstacles represented as red circles. The rectangles shows movements restrictions from \Cref{eq:ding_restriction}, the orange portion shows restricted movement while the green shows where the robot can still move. The orange arrow shows the scaled EE velocity introduced in \Cref{eq:velocity_scaling}, the black arrow is the initial velocity.}
\label{fig:escobedo_controller}\vspace{-16pt}
\end{figure}
This controller's main task equation consists of a third term added to \Cref{eq:ding_avoid_singularities}: 
\begin{equation} \label{eq:hiro_standard_minimization}\small
\begin{split}
g(\mathbf{\dot{q})} & = \frac{1}{2}\mathbf{(\dot{x}_d - J\dot{q})^{\top}(\dot{x}_d - J\dot{q}) +} \frac{\mu}{2}\mathbf{\dot{q}^{\top}\dot{q}}\\
  &\;\;\;\;  + \frac{k}{2}(\mathbf{\dot{q}_{mid}} - \mathbf{\dot{q}})^{\top}(\mathbf{\dot{q}_{mid}} - \mathbf{\dot{q}}).
\end{split}
\end{equation}
The additional third term 
causes the robot to favor joint positions in the middle of its joint limits, keeping the robot away from undesirable joint configurations that may lead to error states.
$k$ is a scaling factor for weighting the middle joint term while $\mathbf{\dot{q}}_{mid}$ represents the desired joint velocity for movement towards the joint's middle position. 

Movement restrictions are utilized to allow the robot to avoid collisions.
As introduced in \Cref{eq:ding_restriction}, $\mathbf{\hat{d}^T J_{c}{\dot{q}}}$ limits the velocity of a particular control point.
The control point's maximum approach velocity $\dot{x}_a$ is computed:
\begin{gather}\small
        \dot{x}_a = \begin{cases}
         V_a- V_{max} : \text{if } d < d_{notice} \text{ and } d < d_{repulse}, \\
       V_b: \text{if } d < d_{notice} \text{ and } d \geq d_{repulse}, \\
        \text{Drop Restriction}: \text{otherwise}, \quad 
        \end{cases}
        \label{eq:approach_velocity_computation} 
\end{gather}
where:
\begin{equation}\small
    V_a = \frac{V_{max}}{1 + e^{\alpha(2\frac{d}{d_{crit}} - 1)}};
\quad
    V_b =  \frac{V_{max}}{1 + e^{\alpha(2\frac{d - d_{crit}}{d_{notice} - d_{crit}} - 1)}}.
    \label{eq:Va_Vb}
\end{equation}
$V_{max}$ is the user-defined maximum magnitude, $d$ is the distance closest point on the robot's body to the object, $d_{repulse}$ is the distance where the repulsive vector reaches a negligible magnitude, and $d_{notice}$ is the distance at which movement restriction begin to be imposed. $d_{crit}$ is the distance at which the the maximum approach velocity $\dot{x}_a$ becomes a negative value, limiting the control point to only move away from an object. Additionally, end-effector velocity is reduced when in proximity to an obstacle:
\begin{equation}\label{eq:velocity_scaling}\small
    \dot{\mathbf{x}}_{\mathbf{d}}=\frac{\left\|d_{\text {lowest }}\right\|}{d_{\max }} \dot{\mathbf{x}}_{\mathbf{d}}
\end{equation}
where $d_{\text {lowest }}$ is the norm of the distance to each obstacle in a region of surveillance within a user-defined maximum distance $d_{\max }$ and selecting the smallest norm, creating a scaling term that reduces the EE velocity. In order to prevent erratic motion from vanishing obstacle readings, a linear decay formula is applied to simulate an obstacle moving away from the robot \cite{escobedo2021contact}.

\section{Evaluation Framework}\label{sec:evaluation}
In this section, the framework used for evaluation of OACs is described. The framework consists of several design considerations and a set of experiments. These design considerations are chosen for their universal application to robot movement analysis. 
\subsection{Design Considerations}

\subsubsection{Kinematics}
Kinematics is a branch of mechanics which deals with the motion of bodies through space and time. Since robots are composed of interconnected bodies and by their very essence designed for motion; kinematics is fundamental for robot analysis. Out of the many aspects of kinematics, we choose to focus on manipulabilty, which is derived from the robot's Jacobian. This metric is used to interpret the capabilities of robot motion at any given time.

Manipulability typically measures the capacity of a robot to position and orient its EE at a given joint configuration. The kinematic chain properties of a robot manipulator allow us to extend this definition from its focus on the EE to any control point. Therefore, manipulability can be used to demonstrate the ability of a control point to move in various directions within the task space, as well as how close it is to a singularity. 

\paragraph{Manipulability ellipsoid}
Using the Jacobian, 
the properties of the manipulator's movement can be examined to produce a metric for assessing a robot's ability to move along the principal axes, as depicted in \Cref{fig:analysis_ellipse_1_high}. This information is extracted from the manipulability ellipsoid. \begin{align}\small
      \mathbf{\dot{q}}^{\top} \mathbf{\dot{q}}=1 \qquad \mathbf{\dot{q}} = \mathbf{J}^{-1} \mathbf{\dot{x}} \qquad  \mathbf{\dot{x}}^{\top} (\mathbf{J} \mathbf{J}^{\top})^{-1} \mathbf{\dot{x}} = 1
      \label{eq:der_ellipsoid1_2}
\end{align}
Consider all EE velocities that can be obtained by choosing joint velocity vectors, $\mathbf{\dot{q}}$, of unit norm. In this case, the left term in \Cref{eq:der_ellipsoid1_2} represents the points on the surface of a sphere in the joint velocity space. Based on the relation of the middle term, the right term represents the points on the surface of an ellipsoid in the EE velocity space.
Where $\mathbf{J J}^{\top}$,  as in Equation \ref{eq:scalar_metric}, defines the properties of a hyperellipsoid in 6D, of which only the 3D translational component will be considered. The geometry of the ellipsoid can be determined by looking at the singular value decomposition of the matrix $\mathbf{J J}^{\top}$, which is a square, real symmetric matrix. The eigenvectors $\boldsymbol{v}$ represent the axial direction of the operability ellipsoid, see \Cref{fig:analysis_ellipse_1_high}, and the square root of the eigenvalues $\lambda$ represent the size of the ellipse with respect to the corresponding axial direction. When the robot reaches a singularity, the Jacobian loses rank and the ellipsoid becomes degenerate. This is characterized by a zero length in one of the principal axis. 
The ellipsoid can be distilled into various metrics that illustrate the controller's impact on the motion of the robot within its manipulability range. For our analysis, we create a metric that tracks the alignment of a repulsive force with the ellipsoid's eigenvector in the direction of lowest axial operability. The metric is defined in \Cref{eq:projection_ellipse} as a scalar projection of the repulsive force $\boldsymbol{V}(\boldsymbol{p})$ from \Cref{eq:repulsive_vector}, or equivalent restriction, on minimum operability eigenvector $\boldsymbol{v_{min}}$. A larger number of occurrences where a control point is directed to move along the axis of least manipulability indicates that a controller is less effective at leveraging kinematics to move the robot away from an obstacle. 
\begin{equation} \label{eq:projection_ellipse}\small
    proj_{\boldsymbol{v_{min}}}\boldsymbol{V}(\boldsymbol{p}) = \frac{\boldsymbol{V}(\boldsymbol{p}) \cdot \boldsymbol{v_{min}}}{{|\boldsymbol{v_{min}}|}^{2}}
\end{equation}

\paragraph{Manipulability scalar:}
A scalar value manipulability metric $w$ is used to represent how freely the robot can move as a relationship between potential movement directions \cite{yoshikawa1985manipulability}. This metric uses the general form of the Jacobian, $\mathbf{J}$, which relates the rate of change of the robot's joint velocities $\mathbf{\dot{q}}$ to the end-effector velocity $\mathbf{\dot{x}}$, as shown in Equation \ref{eq:main_jacobian}.
\begin{equation} \label{eq:scalar_metric}\small
    w = \sqrt{det(\mathbf{J} \mathbf{J}^{\top})} 
\end{equation}
\noindent This measure varies depending on the robot configuration. Larger values of $w$ indicate more freedom of movement in Cartesian space, while smaller values indicate that the robot is approaching a singular configuration. In our experimental scenarios, $w$ is tracked over time as the controller attempts to avoid obstacles in its environment. We evaluate each controller based on its ability to direct the control points away from singularities and towards kinematic regions of high manipulability, measured by a high manipulability scalar.

\subsubsection{Motion Profiles}
A motion profile includes the position, velocity, acceleration, and jerk of the robot EE or joints at every time-step. Trajectory planners often consider motion profiles when describing how a robot should move to a desired position. In the planning process, considerable emphasis is placed on generating smooth motions \cite{haschke2008line}. With the outlined design considerations; it is argued that a controller should generate smooth motion profiles that remain within the bounds of its operational properties, as well as actively responding to dynamic obstacles.

The mechanical properties of a robot pose limit on the range of possible values that compose the motion profile. Acceleration, deceleration, and velocity are limited by the inertia of the mechanical system, while the cartesian path of the robot links is confined to its task space. Jerk must remain within the robot's joint limits because the current of the motor cannot be instantly changed. Commands to the motors must also stay within the bounds introduced by the The International Organisation for Standardization (ISO), which restrict allowable power and force as a measure for implementing mechanically safe collaborative robots (cobots) \cite{ISO2016cobot}.

In this work, emphasis is placed on the jerk profiles of the joints. Minimizing jerk reduces vibration, wear on mechanical parts, and the potential for servo errors \cite{vass2003real}. To assess this aspect of the robot's movement, the jerk profiles are calculated over the course of trajectory execution and obstacle avoidance. This data is used as a metric of comparison in the analysis of OACs. Spikes in jerk demonstrate a controller's inability to provide smooth velocity changes to a robot during obstacle avoidance. 
\subsubsection{Virtual Constraints}
Through analysis of virtual constraints, the design principles of various OACs can be verified. Virtual constraints represent relations between the state variables of a mechanical model. OACs use forces at control points to form virtual constraints. Measuring the properties of repulsive forces and control points demonstrates how obstacle inputs trigger reactive behaviors. The magnitude of the force, for example, can be used to assess the reactivity of a controller when acting on a given control point. The location of a control point can tell us which part of the robot body is used as a reference for obstacle behavior. Different parts of the robot's body have varying properties, such as manipulability, as discussed in previous sections. In this framework, the shape of the repulsive force profile in relation to the minimum distance to an obstacle is used as a metric for virtual constraints. The increase or decrease of the repulsive force at every point in time should taken into consideration for analysis of each controller defined in \Cref{sec:methods}.

\subsection{Experiments}\label{sec:methods:experimental_setp}
An overarching purpose of our framework is the experimental evaluation of obstacle avoidance behaviors of OACs in various obstacle-robot interaction scenarios. Examples  of possible variation among experiments include the number of obstacles, velocity of obstacles, direction of obstacle travel, trajectory of robot, and point of contact between the robot and an obstacle. To demonstrate the merit of our framework, we conduct the following experiments with three OACs, described in \Cref{sec:methods}.

All experiments are conducted on a real 7-Degree of Freedom (7-DoF) Franka Emika Panda robotic arm. 
Controllers are developed using C++ and ROS.
Objects are introduced into the robot's environment virtually, using rostopics to guarantee that each controller has access to the exact same information.

\subsubsection{Static Robot, Dynamic Obstacle (SRDO)}\label{sec:r-static-o-dynamic}

In this experiment. the manipulator is commanded to maintain a static end-effector position $p_{EE}$ = $(x_{EE}, y_{EE}, z_{EE})$ = $(0.4, 0.0, 0.45)[m]$ from the robot's base. An obstacle is presented moving towards the robot body at a rate of $0.15$ m/s. The obstacle starts at $(0, -0.5, 0.6)$ and ends at $(0, 0.1, 0.6)$.

\subsubsection{Dynamic Robot, Dynamic Obstacle (DRDO)}

In this scenario, the manipulator is commanded to move in a Cartesian circle with a radius of 0.25 m centered at $(0.5, 0, 0.25)$ moving counter-clockwise in the $x$ and $y$ directions while an object moves near the EE.
The robot moves at a maximum speed of $0.3$ m/s when there are no obstacles in the vicinity. To trigger avoidance behaviors, an obstacle moves from $(0.45, -0.5, 0.45)$ to $(0.45, 0.1, 0.45)$ at $0.15$ m/s.











%

%
\section{Results \& Discussion}\label{sec:results_discussion}
In the following sections, we summarize the results obtained after running the experiments in \Cref{sec:methods:experimental_setp}. We demonstrate that we can effectively evaluate OACs based on two simple experiments and four separate metrics, which originate from our design considerations. We compare the controllers based on these performance metrics and make suggestions for directions of future work. Additional analysis and real-time operation is detailed in the accompanying video.
\subsection{Manipulability scalar}
\begin{figure} \centering
    \begin{subfigure}{0.24\textwidth} 
        \includegraphics[width=0.99\textwidth]{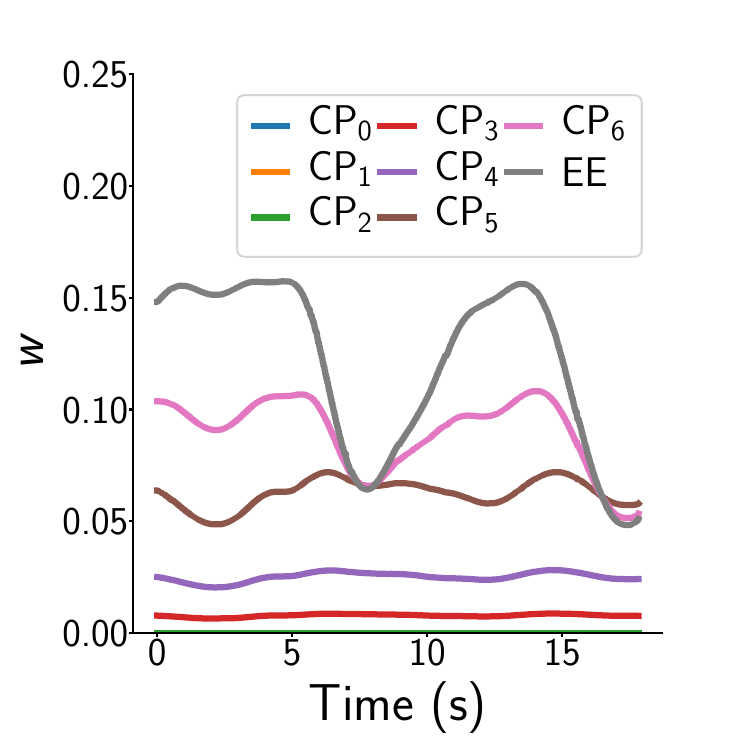}
        \label{fig:flacco-circle_line_to_ee_manip}
    \end{subfigure} \hspace{-1em}
    \begin{subfigure}{0.24\textwidth}
        \includegraphics[width=0.99\textwidth]{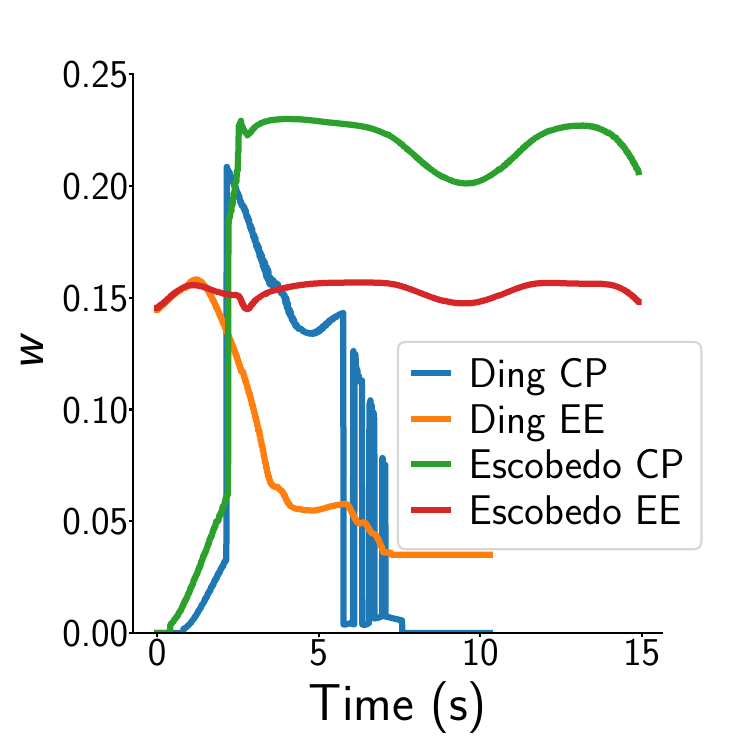}
        \label{fig:ding-circle_line_to_ee_manip}
    \end{subfigure}
    \vspace{-1\baselineskip}
  \caption{The manipulability at the control points (CP) and the end-effector (EE) tracked across time for the five control points of Flacco's work (left), and the control point and end-effector of Ding and Escobedo (right) Both graphs are constructed from the DRDO experimental scenario.}
  \label{fig:circle_line_to_ee_manip}
  \vspace{-12pt}
\end{figure}

\begin{figure} \centering
    \begin{subfigure}{0.24\textwidth}
        \includegraphics[trim=80 0 0 0, clip, width=0.99\textwidth]{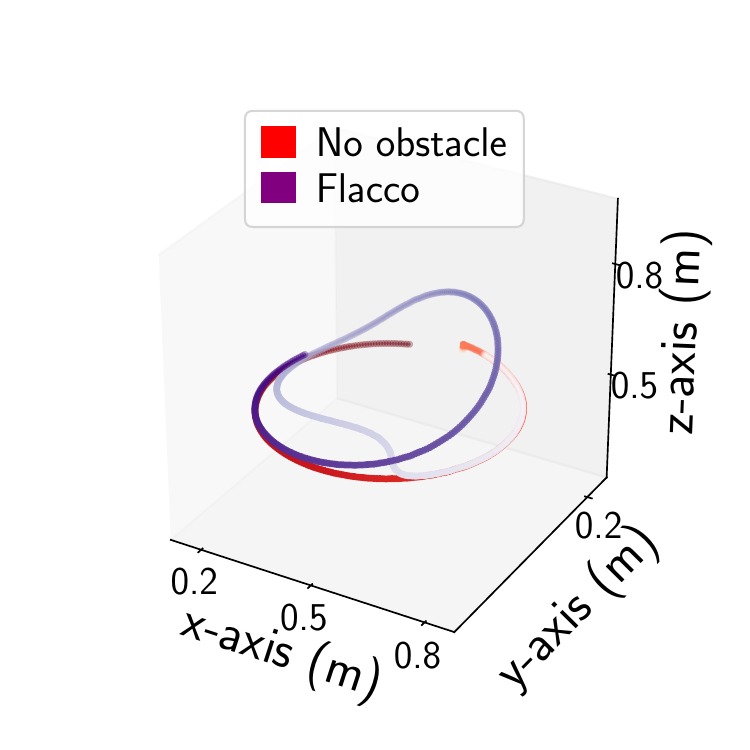}
        \label{fig:none-flacco-circle_line_to_ee_traj}
    \end{subfigure} \hspace{-1em}
    \begin{subfigure}{0.24\textwidth} 
        \includegraphics[trim=80 0 0 0, clip, width=0.99\textwidth]{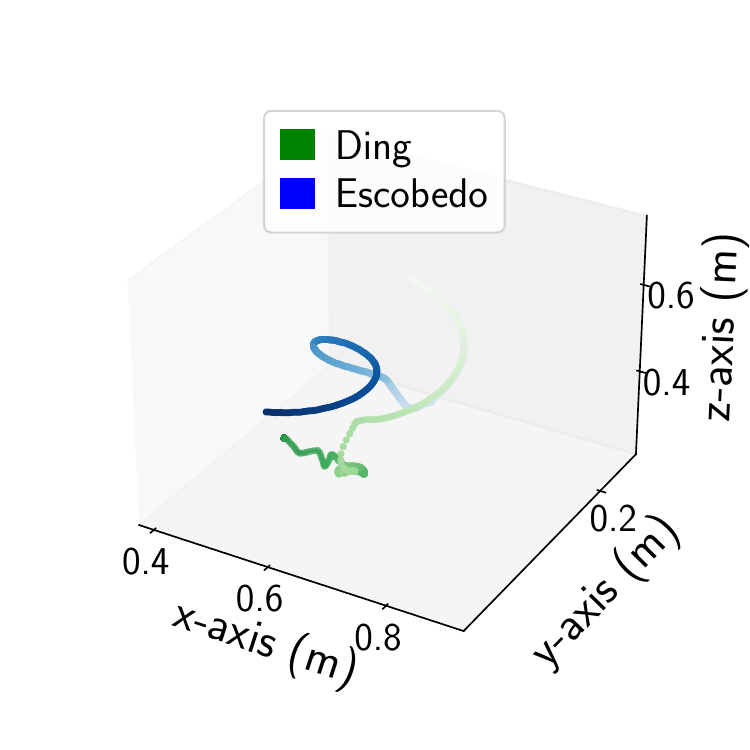}
        \label{fig:ding-hiro-circle_line_to_ee_traj}
    \end{subfigure}
 
  \caption{The end-effector path of the robot as it is commanded by each controller to travel in a circular path while being approached by an obstacle. Both graphs are constructed from the DRDO experimental scenario.}
  \label{fig:circle_line_to_ee_traj}
  \vspace{-12pt}
\end{figure}

The manipulability values for control points and EE are shown in \Cref{fig:circle_line_to_ee_manip} for the dynamic robot experiment. 
The manipulability scalar values of Flacco's control points are smooth in both scenarios, as opposed to Ding and Escobedo. We see an equivalent pattern for the static robot experiment. 
In \Cref{fig:circle_line_to_ee_manip} we can see that both Ding and Escobedo have sudden changes in manipulability as the obstacle point moves closer to the robot's body. This is the result of a rapid change of the control point position and a consequent change of the movement restriction to a new part of the robot's body. 
Any discontinuity of control points can lead to undesirable movements, as can be seen in the EE path for Ding in \Cref{fig:circle_line_to_ee_traj}. Not only is the discontinuity sudden, but the transition is between two points with drastically different manipulability measures. 
Importantly, these points should not be treated equally when applying movement restrictions: a point with higher manipulability can likely avoid an obstacle without effecting the main task. 
We conclude that information about the current control point's manipulability measure should be used to ensure that a repulsive force is scaled accordingly. 
\subsection{Manipulability ellipsoid}
\begin{figure} \centering

    \begin{subfigure}{0.24\textwidth}
        \includegraphics[width=0.99\textwidth]{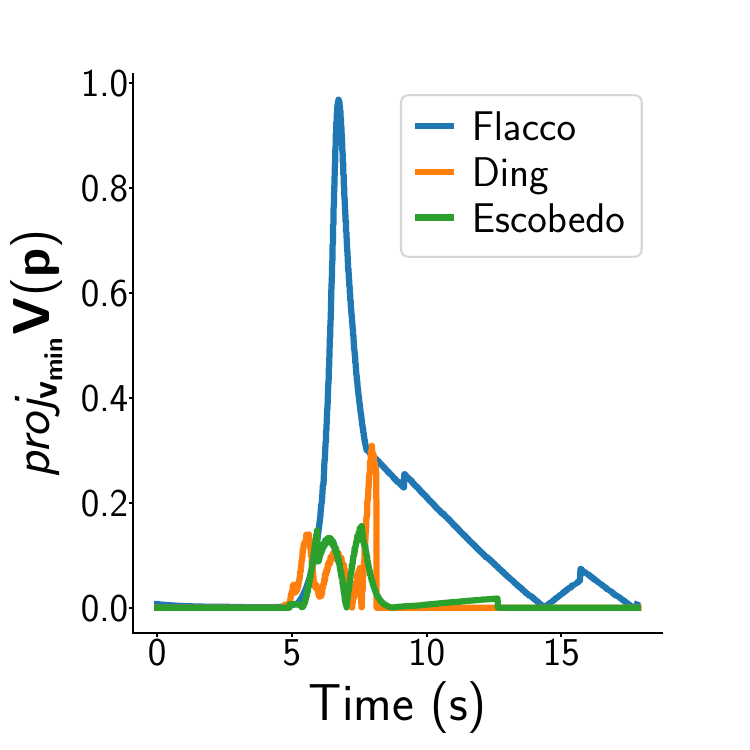}
    \label{fig:all-static_line_to_body_proj}
    \end{subfigure} \hspace{-1em}
    \begin{subfigure}{0.24\textwidth} 
        \includegraphics[width=0.99\textwidth]{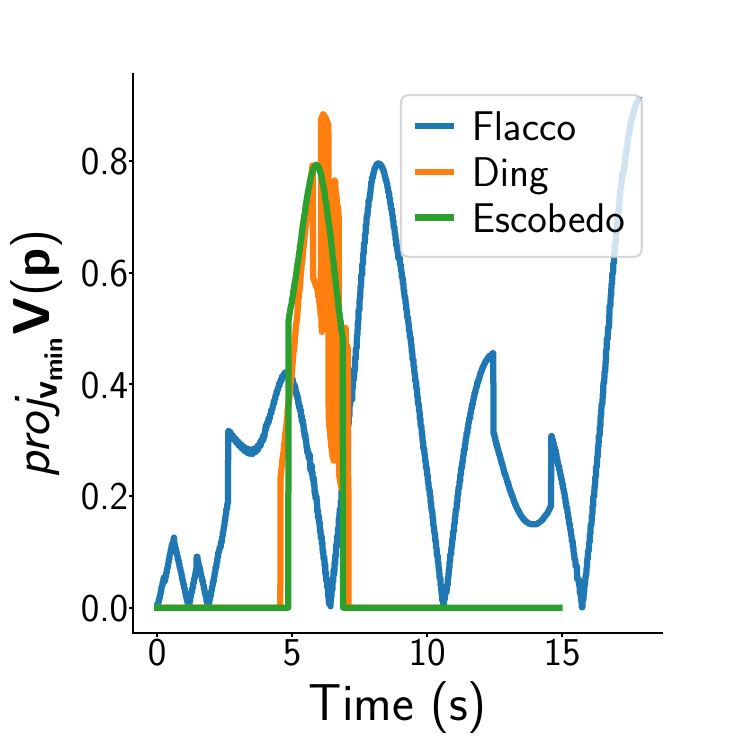}
        \label{fig:all-circle_line_to_ee_proj}
    \end{subfigure}
 \vspace{-1\baselineskip}
  \caption{The proportion of the repulsive vector's magnitude projected onto the minimum dimension of the manipulability ellipsoid as per \Cref{eq:projection_ellipse}. In the left graph is from the SRDO scenario and the in the right graph is from the DRDO scenario.}
  \label{fig:ellipsoid-proj-all}
    \vspace{-14pt}
\end{figure}

\Cref{fig:ellipsoid-proj-all} shows the amount that each controller's repulsive vector aligns with the direction of the control point's lowest axial operability as calculated in \Cref{eq:projection_ellipse}. This metric is not directly taken into account for EE or control point motion within any of the controllers, and is therefore useful to compare how the robot's ability to move is affected in a configuration resulting from avoiding an object. 
Movement meant to avoid a collision can lead to undesirable robot configurations that cannot be recovered from when an object is present, because the control points are commanded to move in a direction that is not feasible in its current kinematic configuration. In \Cref{fig:ellipsoid-proj-all} when the majority of the movement required to avoid a collision is along the axis of minimum manipulability, we see a value near one. This indicates that the control point is being commanded to move in the direction of lowest manipulability.
While any control point can have a low manipulability measure in a particular configuration, control points near the base of the robot manipulator will always have low manipulability because they are kinematically more constrained, as shown in \Cref{fig:circle_line_to_ee_manip}. These control point locations can be seen in \Cref{fig:rviz}.
Therefore, imposing constraints on the proximal links of the manipulator naturally causes a mismatch between the desired and achievable velocity. In other words, the robot cannot physically move the proximal-link control points beyond the manipulability defined by the ellipsoid. A lack of motion at the control points can inhibit the main objective function defined by \Cref{eq:standard_minimization} and lead to an unsolvable optimization formulation. To avoid unrealistic constraints, control points and restrictions should not be placed near the base of the manipulator.
\subsection{Motion Profiles}
\begin{figure} \centering
    \begin{subfigure}{0.23\textwidth}
        \includegraphics[width=0.99\textwidth]{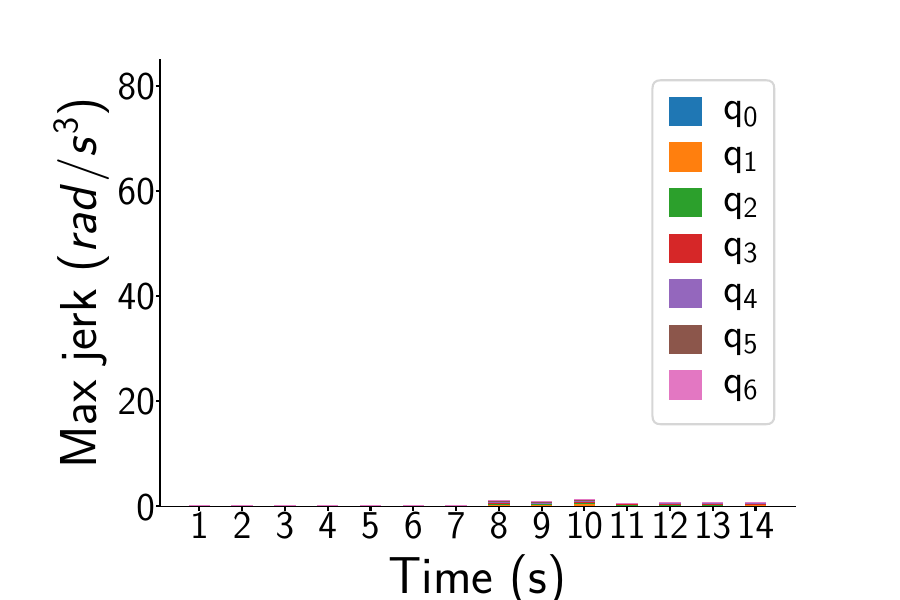}
        \caption{No obstacle}
        \label{fig:none-circle_line_to_ee_jerk}
    \end{subfigure}
    \begin{subfigure}{0.23\textwidth} 
        \includegraphics[width=0.99\textwidth]{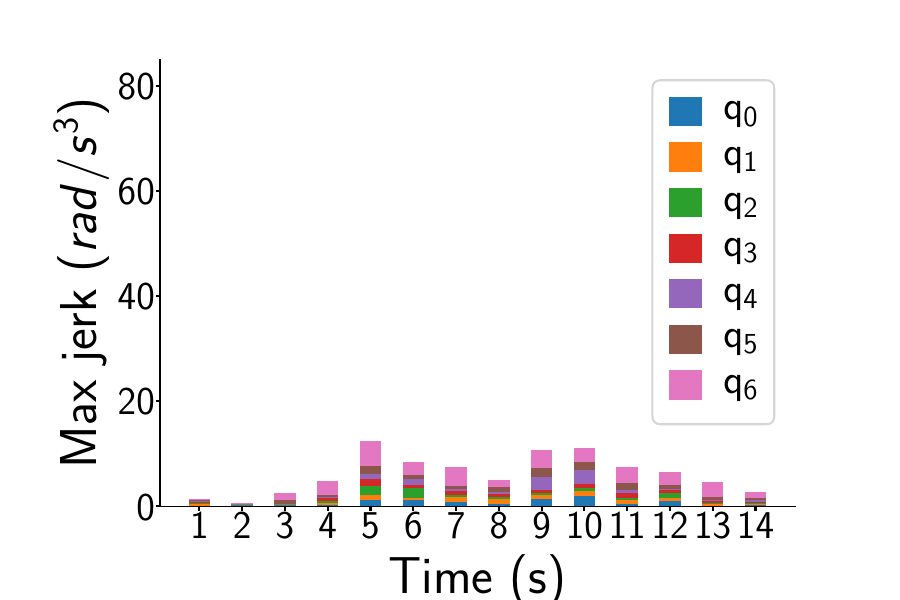}
        \caption{Flacco}
        \label{fig:flacco-circle_line_to_ee_jerk}
    \end{subfigure}\hfill
    \begin{subfigure}{0.23\textwidth}
        \includegraphics[width=0.99\textwidth]{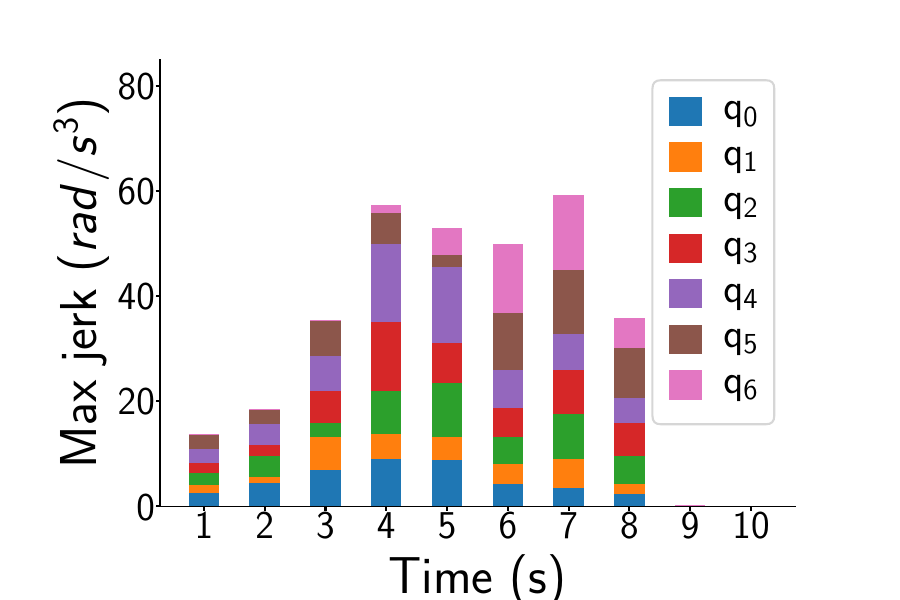}
        \caption{Ding}
        \label{fig:ding-circle_line_to_ee_jerk}
    \end{subfigure}
    \begin{subfigure}{0.23\textwidth}
        \includegraphics[width=0.99\textwidth]{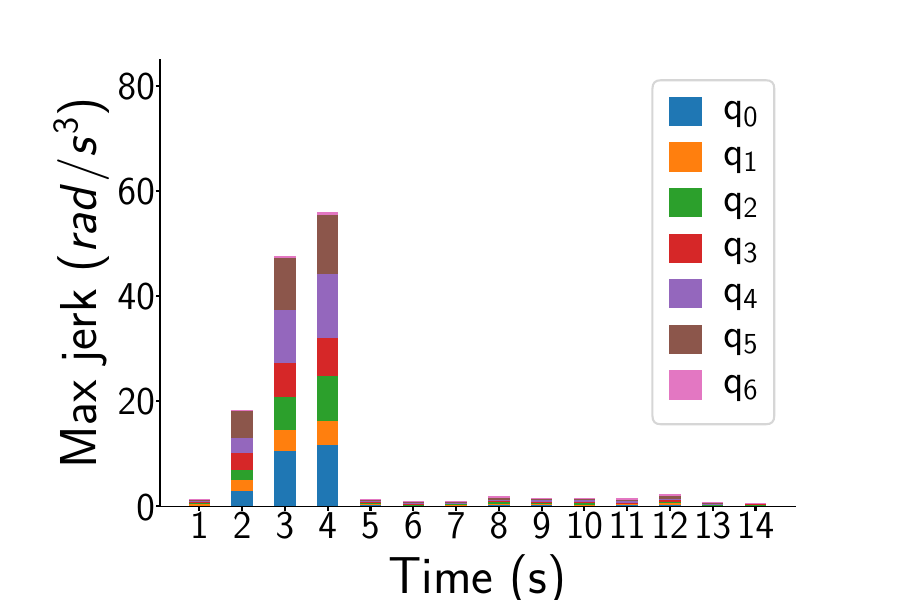}
        \caption{Escobedo}
        \label{fig:hiro-circle_line_to_ee_jerk}
    \end{subfigure}
  \caption{Maximum jerk per second of experimental time plotted for each joint of the robot arm. The spikes in jerk correspond to avoidance behaviors in the DRDO scenario. (a) uses \Cref{eq:ding_avoid_singularities} with no added avoidance terms to follow the specified trajectory.}
  \label{fig:circle_line_to_ee_jerk}
  \vspace{-14pt}
\end{figure}

The jerk profiles capture how smoothly the robot moves through space. \Cref{fig:circle_line_to_ee_jerk} shows the maximum jerk as the robot moves in a circular path and an obstacle approaches the EE. 
Without the obstacle, the robot's acceleration does not rapidl change. However, when approached by an obstacle, the controllers impose a virtual force, altering the velocity. These changes in velocity are apparent through jerk.
The measured discrete velocity is read from the Franka Panda API and smoothed using a Savitzky-Golay filter to increase the precision of the data without distorting the signal tendency \cite{savitzky1964smoothing}. The gradient is then computed using second order accurate central differences in the interior points. We calculate the gradient twice to approximate jerk from the velocity profiles.
In \Cref{fig:none-circle_line_to_ee_jerk}, the ``no obstacle'' graph incurs non-zero jerk, which is likely the baseline jerk of the executed trajectory. 
Flacco (\Cref{fig:flacco-circle_line_to_ee_jerk}) successfully avoids the obstacle despite showing slight increases in jerk during avoidance behavior.
Ding (\Cref{fig:ding-circle_line_to_ee_jerk}) ran for 8 seconds until the joints contorted into an undesirable configuration and the robot entered a self-collision state.
Escobedo (\Cref{fig:hiro-circle_line_to_ee_jerk}) showed low jerk after 4 seconds because it slowed to a stop by design as the obstacle approached the end-effector.
Based on \Cref{fig:circle_line_to_ee_jerk}, it is apparent that the restrictions placed on Ding and Escobedo do not result in smooth avoidance behavior. Each controller's jerk profile is unique in the joints that exhibit jerky behavior. Comparatively, the joint jerk for Escobedo $\mathbf{q}_6$ is less pronounced than both Flacco and Ding.
These jerk profile metrics indicate that adjusting the applied forces in \Cref{eq:ding_restriction} and \Cref{eq:approach_velocity_computation} is required to smooth the robot motion profile. 
Furthermore, low jerk is synonymous with low motor current, which must be required to meet ISO safety standards for robot operation \cite{ISO2016cobot}.
\subsection{Virtual Constraints}
\begin{figure} \centering
    \includegraphics[width=0.48\textwidth]{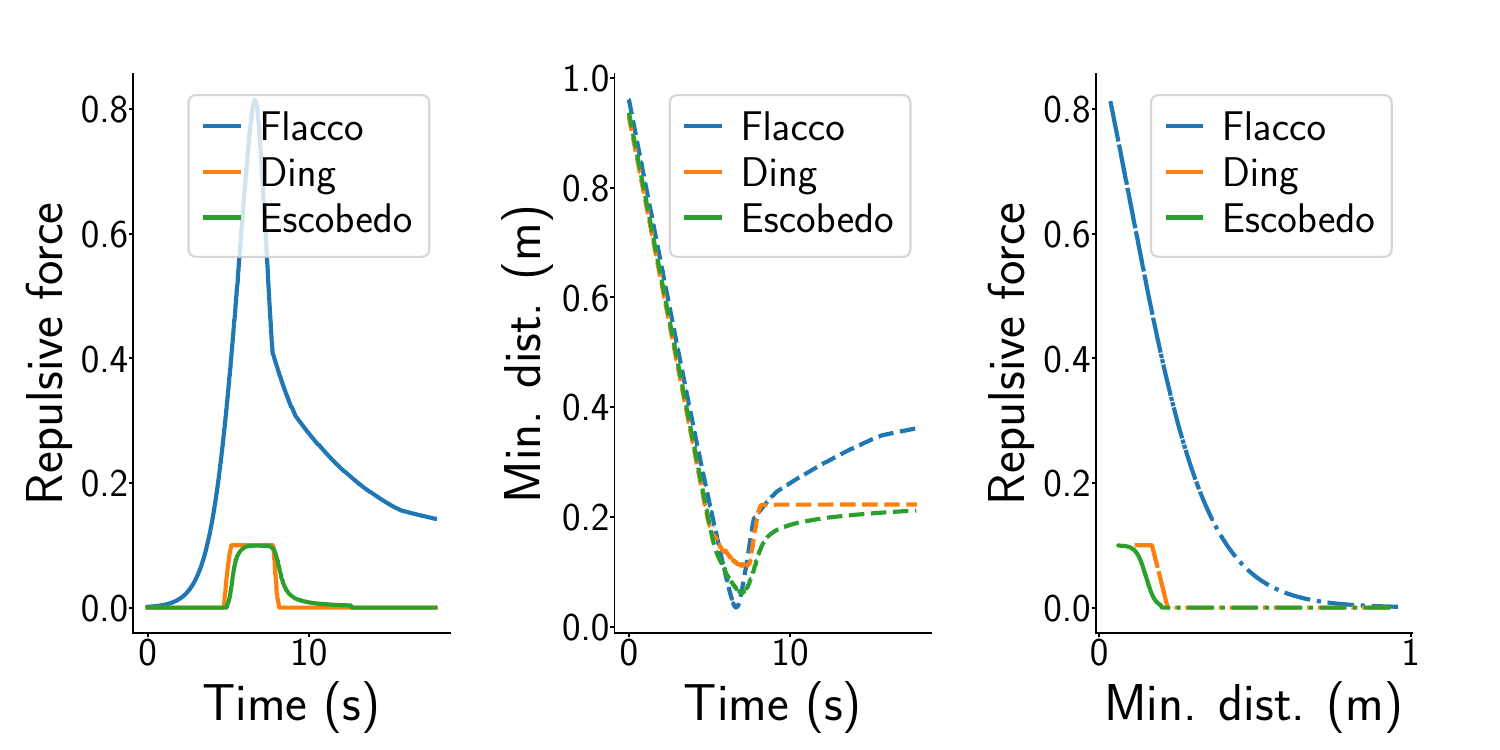}
    \caption{Repulsive vector applied at the control point over time (left). Minimum distance from obstacle to the control point overtime (center). Minimum distance from the obstacle to the control point over the repulsive force applied to the control point (right). Data taken from the SRDO scenario.}
    \label{fig:all-static_line_to_body_repulse}
    \vspace{-12pt}
\end{figure}

In \Cref{fig:all-static_line_to_body_repulse} we plot the virtual repulsive force and minimum distance from obstacle to control point. The repulsive forces are generated from each of the controller's respective avoidance equations. The repulsive forces in these graphs should be designed and shaped for a particular usage and verified in simple experimental scenarios, such as those introduced in this work. For example, in \Cref{fig:all-static_line_to_body_repulse}, Flacco sees a sharp rise in repulsive force at 0.3 meters distance. This force quickly reaches its maximum value which causes the robot to quickly move away from an obstacle. While Flacco will try to avoid nearby obstacles, that is not always the goal when interacting around humans. In Escobedo, when an obstacle approaches the EE, the repulsive force is mitigated by the EE velocity scaling introduced in \Cref{eq:velocity_scaling}. This particular behavior was designed to avoid high velocities when a human is interacting with a robot.
\section{Conclusions}\label{sec:conclusions}
This paper presents an evaluation framework for object-aware controllers (OACs). Three design considerations (kinematics, motion profiles, and virtual constraints) are used as assessment metrics in a set of experimental scenarios where an obstacle enters the robot's workspace and alters the robot's path. Three representative OACs are evaluated on a real robot with dynamic virtual obstacles. 
It is demonstrated that, by using our framework, we can see the following shortcomings of the implemented OACs based on their experimental comparison: First, OACs fail to effectively incorporate kinematic information about the specific robot embodiment as part of their movement. Without kinematic properties, such as manipulability, the robot can get trapped in unrealistic constraints that impede its ability to function. 
Second, OACs set discontinuous virtual constraints which lead to unsteady motion profiles, which are highly undesirable for general robot safety.
Finally, OACs specify unweighted constraints, which fail to prioritize robot behaviors in situations where multiple objectives need to be achieved. We conclude that using our framework to find a measurable and testable interpretation of OACs is essential to their improvement and accessibility.
Matching measurable properties of OACs with robot behavior advises future directions of research and provides more confidence in adoption of these methods to ensure robot safety. In the future, we envision the expansion of this framework by the research community to build a library of controller comparison benchmarks. 

\addtolength{\textheight}{-4cm}
\bibliographystyle{IEEEtran}
\bibliography{root}

\begin{thebibliography}{10}
\providecommand{\url}[1]{#1}
\csname url@rmstyle\endcsname
\providecommand{\newblock}{\relax}
\providecommand{\bibinfo}[2]{#2}
\providecommand\BIBentrySTDinterwordspacing{\spaceskip=0pt\relax}
\providecommand\BIBentryALTinterwordstretchfactor{4}
\providecommand\BIBentryALTinterwordspacing{\spaceskip=\fontdimen2\font plus
\BIBentryALTinterwordstretchfactor\fontdimen3\font minus
  \fontdimen4\font\relax}
\providecommand\BIBforeignlanguage[2]{{%
\expandafter\ifx\csname l@#1\endcsname\relax
\typeout{** WARNING: IEEEtran.bst: No hyphenation pattern has been}%
\typeout{** loaded for the language `#1'. Using the pattern for}%
\typeout{** the default language instead.}%
\else
\language=\csname l@#1\endcsname
\fi
#2}}

\bibitem{zacharaki2020safety}
A.~Zacharaki, I.~Kostavelis, A.~Gasteratos, and I.~Dokas, ``Safety bounds in
  human robot interaction: A survey,'' \emph{Safety science}, vol. 127, p.
  104667, 2020.

\bibitem{lasota2017survey}
P.~A. Lasota, T.~Fong, J.~A. Shah, \emph{et~al.}, \emph{A survey of methods for
  safe human-robot interaction}.\hskip 1em plus 0.5em minus 0.4em\relax Now
  Publishers, 2017.

\bibitem{flacco2015control}
F.~Flacco, A.~De~Luca, and O.~Khatib, ``Control of redundant robots under hard
  joint constraints: Saturation in the null space,'' \emph{IEEE Transactions on
  Robotics}, vol.~31, no.~3, pp. 637--654, 2015.

\bibitem{ding2020collision}
Y.~Ding and U.~Thomas, ``Collision avoidance with proximity servoing for
  redundant serial robot manipulators,'' in \emph{2020 IEEE International
  Conference on Robotics and Automation (ICRA)}.\hskip 1em plus 0.5em minus
  0.4em\relax IEEE, 2020.

\bibitem{escobedo2021contact}
C.~Escobedo, M.~Strong, M.~West, A.~Aramburu, and A.~Roncone, ``Contact
  anticipation for physical human--robot interaction with robotic manipulators
  using onboard proximity sensors,'' in \emph{IEEE/RSJ International Conference
  on Intelligent Robots and Systems (IROS)}, 2021.

\bibitem{haddadin2017robot}
S.~Haddadin, A.~De~Luca, and A.~Albu-Sch{\"a}ffer, ``Robot collisions: A survey
  on detection, isolation, and identification,'' \emph{IEEE Transactions on
  Robotics}, vol.~33, no.~6, pp. 1292--1312, 2017.

\bibitem{flacco2012depth}
F.~Flacco, T.~Kr{\"o}ger, A.~De~Luca, and O.~Khatib, ``A depth space approach
  to human-robot collision avoidance,'' in \emph{2012 IEEE International
  Conference on Robotics and Automation}.\hskip 1em plus 0.5em minus
  0.4em\relax IEEE, 2012.

\bibitem{rakita2021collisionik}
D.~Rakita, H.~Shi, B.~Mutlu, and M.~Gleicher, ``Collisionik: A per-instant pose
  optimization method for generating robot motions with environment collision
  avoidance,'' in \emph{2021 IEEE International Conference on Robotics and
  Automation (ICRA)}.\hskip 1em plus 0.5em minus 0.4em\relax IEEE, 2021, pp.
  9995--10\,001.

\bibitem{merckaert2022real}
K.~Merckaert, B.~Convens, C.-j. Wu, A.~Roncone, M.~M. Nicotra, and
  B.~Vanderborght, ``Real-time motion control of robotic manipulators for safe
  human--robot coexistence,'' \emph{Robotics and Computer-Integrated
  Manufacturing}, vol.~73, p. 102223, 2022.

\bibitem{fan2021aurasense}
X.~Fan, R.~Simmons-Edler, D.~Lee, L.~Jackel, R.~Howard, and D.~Lee,
  ``Aurasense: Robot collision avoidance by full surface proximity detection,''
  in \emph{2021 IEEE/RSJ International Conference on Intelligent Robots and
  Systems (IROS)}.\hskip 1em plus 0.5em minus 0.4em\relax IEEE, 2021, pp.
  1763--1770.

\bibitem{nguyen2018compact}
D.~H.~P. Nguyen, M.~Hoffmann, A.~Roncone, U.~Pattacini, and G.~Metta, ``Compact
  real-time avoidance on a humanoid robot for human-robot interaction,'' in
  \emph{2018 13th ACM/IEEE International Conference on Human-Robot Interaction
  (HRI)}.\hskip 1em plus 0.5em minus 0.4em\relax IEEE, 2018, pp. 416--424.

\bibitem{roncone2015learning}
A.~Roncone, M.~Hoffmann, U.~Pattacini, and G.~Metta, ``Learning peripersonal
  space representation through artificial skin for avoidance and reaching with
  whole body surface,'' in \emph{2015 IEEE/RSJ International Conference on
  Intelligent Robots and Systems (IROS)}.\hskip 1em plus 0.5em minus
  0.4em\relax IEEE, 2015, pp. 3366--3373.

\bibitem{khatib1986real}
O.~Khatib, ``Real-time obstacle avoidance for manipulators and mobile robots,''
  in \emph{Autonomous robot vehicles}.\hskip 1em plus 0.5em minus 0.4em\relax
  Springer, 1986, pp. 396--404.

\bibitem{ding2021improving}
Y.~Ding and U.~Thomas, ``Improving safety and accuracy of impedance controlled
  robot manipulators with proximity perception and proactive impact
  reactions,'' in \emph{2021 IEEE International Conference on Robotics and
  Automation (ICRA)}.\hskip 1em plus 0.5em minus 0.4em\relax IEEE, 2021, pp.
  3816--3821.

\bibitem{rakita2018relaxedik}
D.~Rakita, B.~Mutlu, and M.~Gleicher, ``Relaxedik: Real-time synthesis of
  accurate and feasible robot arm motion.''

\bibitem{Nakamura1986}
Y.~Nakamura and H.~Hanafus, ``Manipulability of robotic mechanisms,'' 1986.

\bibitem{yoshikawa1985manipulability}
T.~Yoshikawa, ``Manipulability of robotic mechanisms,'' \emph{The International
  Journal of Robotics Research}, vol.~4, no.~2, pp. 3--9, 1985.

\bibitem{haschke2008line}
R.~Haschke, E.~Weitnauer, and H.~Ritter, ``On-line planning of time-optimal,
  jerk-limited trajectories,'' in \emph{2008 IEEE/RSJ International Conference
  on Intelligent Robots and Systems}.\hskip 1em plus 0.5em minus 0.4em\relax
  IEEE, 2008, pp. 3248--3253.

\bibitem{ISO2016cobot}
``{Robots and robotic devices — Collaborative robots},'' International
  Organization for Standardization, Geneva, CH, Standard ISO/TS 15066:2016,
  Feb. 2016.

\bibitem{vass2003real}
G.~Vass, B.~Lantos, and S.~Payandeh, ``Real-time optimized robot trajectory
  planning with jerk,'' \emph{IFAC Proceedings Volumes}, vol.~36, no.~17, pp.
  265--270, 2003.

\bibitem{savitzky1964smoothing}
A.~Savitzky and M.~J. Golay, ``Smoothing and differentiation of data by
  simplified least squares procedures.'' \emph{Analytical chemistry}, vol.~36,
  no.~8, pp. 1627--1639, 1964.

\end{thebibliography}
\end{document}